%% file: root.tex
\documentclass[letterpaper,twocolumn,fleqn]{article} 
\usepackage{ist}
\usepackage{cite}
\usepackage{amsmath,amssymb,amsfonts}
\usepackage{algorithmic}
\usepackage{graphicx}
\usepackage{textcomp}
\usepackage{xcolor}
\usepackage{url}
\usepackage{float}
\usepackage{subcaption}
\usepackage{diagbox}
\usepackage{graphicx}
\usepackage{multirow}
\usepackage{adjustbox} 
\def\BibTeX{{\rm B\kern-.05em{\sc i\kern-.025em b}\kern-.08em
    T\kern-.1667em\lower.7ex\hbox{E}\kern-.125emX}}

\title{Drone Object Detection Using RGB/IR Fusion\\
}

\author{Lizhi Yang, Ruhang Ma, Avideh Zakhor \newline \{lzyang, m.ruhang, avz\}@berkeley.edu  \newline Department of Electrical Engineering and Computer Science\newline University of California, Berkeley
}
\begin{document}
\maketitle

\begin{abstract}
Object detection using aerial drone imagery has received a great deal of attention in recent years. While visible light images are adequate for detecting objects in most scenarios, thermal cameras can extend the capabilities of object detection to night-time or occluded objects. As such, RGB and Infrared (IR) fusion methods for object detection are useful and important. One of the biggest challenges in applying deep learning methods to RGB/IR object detection is the lack of available training data for drone IR imagery, especially at night. In this paper, we develop several strategies for creating synthetic IR images using the AIRSim simulation engine and CycleGAN. Furthermore, we utilize an illumination-aware fusion framework to fuse RGB and IR images for object detection on the ground. We characterize and test our methods for both simulated and actual data. Our solution is implemented on an NVIDIA Jetson Xavier running on an actual drone, requiring about 28 milliseconds of processing per RGB/IR image pair.

\end{abstract}

\input{introduction}
\input{method}

\input{experiments}
\input{conclusion}

\begin{biography}
Lizhi Yang is currently a senior undergraduate student at UC Berkeley. His work 
focuses on legged robotics, human-robot interaction, and computer vision.

Ruhang Ma is a graudated student of UC Berkeley. His work focuses on computer vision.

Avideh Zakhor is currently Qualcomm Chair and professor in EECS at U.C. Berkeley. Her areas of interest include theories and applications of signal, image and video processing and 3D computer vision. She has won a number of best paper awards, including the IEEE Signal Processing Society in 1997 and 2009, IEEE Circuits and Systems Society in 1997 and 1999 and IEEE Solid Circuits Society in 2008.
\end{biography}

\end{document}

%% file: introduction.tex
\section{1. Introduction}
Object detection has proven to be useful in many fields, be it cable inspection, livestock monitoring or general surveillance. 
Compared with land-based methods, using a drone as a platform for the payload has the advantage of being able to traverse any terrain in a stable, predictable fashion. 
However, current drone object detection methods use the onboard visible light RGB camera, which are only suitable for daylight capture times, and necessarily perform poorly during night settings. 
Thus, it is desirable to develop new methods of object detection that are invariant to the lighting conditions of the environment.  
Infrared (IR) imagery is an obvious choice for object detection at night-time; however, the performance of infrared imagery in daytime is typically lower than RGB imagery since most RGB cameras have considerably higher pixel count and pixel resolution than IR cameras. 
As such, it is natural to combine IR and RGB imagery for object detection via an adaptive illumination aware network (IAN) \cite{guan-illumination-aware,li-illumination-aware}, to achieve the best of both worlds at day and night-time.

To apply deep learning methodologies to solve this problem, we need pairs of IR/RGB imagery to train deep networks.
Although there exist datasets for drone object detection, for the most part, they are either RGB \cite{zhu2018vision,zhu2020vision}or IR \cite{iricra2014}. 
The few dataset with both IR and RGB imagery either do not have registered frames \cite{leykin2007thermal}, or they are not captured by drones \cite{takumi2017multispectral,coaxials-2019}. 
In addition, existing datasets are all captured in urban areas, which could be limiting in more general settings. 
Finally, many recent works on fusing RGB and IR channels propose application specific deep networks for applications such as pedestrian detection, leak checking or autonomous driving \cite{multispectral,guan-illumination-aware,li-illumination-aware,leakage}, which are not necessarily applicable to drone captured data.  

In this paper, we describe three approaches to overcome lack of availability of paired RGB/IR images with labels for training purposes.  
The first one uses a CycleGAN \cite{CycleGAN2017} which aims to synthetically generate corresponding IR images once it is presented with RGB images.
Since there is an abundance of labeled RGB drone imagery, this process can result in labeled IR images.
Even if labeled drone imagery were not readily available, there exist accurate RGB object detectors for drone imagery \cite{zhu2020vision}, which can be applied to label the RGB images with relatively high accuracy e.g., above 90\%. 
However, CycleGAN alone cannot recover lost information in RGB images taken at night time.
Therefore it is not possible for a synthetic IR image trained with RGB imagery to reveal information that would have been present in an actual IR image taken at night time with a thermal camera.  

To this end, we employ an open-source simulation environment Aerial Informatics and Robotics Simulation (AIRSim) \cite{shah2017airsim,bondi2018airsim} to synthesize realistic IR images at day or night time.
AIRSim provides simulation for drones, ground vehicles such as cars and various other objects, built on Unreal Engine 4 \cite{unreal} to create physically and visually realistic simulations. 
We use AIRSim to create synthetic environments, not necessarily limited to urban environments, to render photorealistic IR/RGB images at day and night time.
Our third approach to creating labeled IR imagery involves applying CycleGAN to the rendered RGB images from the AIRSim simulator  to close the sim-to-real gap between simulation and the real world. 
We characterize the performance of all the three synthetic IR generation methods in the context of IR/RGB fusion for object detection in drone captured imagery.

The outline of this paper is as follows: In Section 2, we discuss the specifics of the RGB/IR fusion model used throughout the paper.
In Section 3, we discuss the three synthetic IR generation methods in detail.
In Section 4, we test our fusion model on both synthetically generated IR data and real-world datasets. 
Conclusions and future work are included in Section 5.

\begin{figure}[ht]
    \centering
    \includegraphics[width=\linewidth]{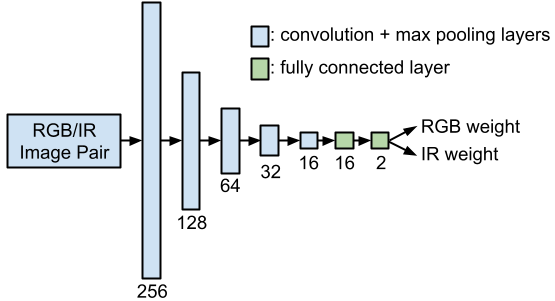}
    \caption{Structure of the Illumination Awareness Network (IAN) comprised of 7 layers with the first 5 being convolution + max pooling layers and the last 2 being fully connected layers. It takes a RGB/IR image pair as input and outputs the RGB and IR weights respectively.}
    \label{fig:IAN}
\end{figure}

%% file: method.tex
\label{sec:method-object-detection}
\section{2. RGB/IR fusion object detection}

In this section we describe our basic design choices and methodology for IR/RGB fusion. We use YOLOv4 \cite{yolov4} as the basic architecture for our RGB and IR object detectors. 
YOLOv4 is a real-time object detection model which has been shown to achieve fast inference time as well as high accuracy. Compared to different object detection methods \cite{FasterR-CNN,RetinaNet,SSD,yolov3,tan2020efficientdet}, its one-stage design avoids computing expensive region proposal network, making it twice as fast as EfficientDet \cite{tan2020efficientdet} with comparable performance on MS COCO dataset \cite{lin2015microsoft-coco}. It also uses Spatial Pyramid Pooling (SPP) \cite{spp} to deal with scale invariance. 

We retrain a YOLOv4 model on the VisDrone dataset \cite{zhu2018vision,zhu2020vision} and use it as our RGB detector in the remainder of the paper. 
VisDrone dataset is collected by drone platforms under various weather and lighting conditions and contains 261,908 frames and 10,209 static images.
Our IR object detection model is constructed by reducing the 3 RGB channels in the original YOLOv4 model to 1 infrared channel.
We train the IR model using synthetic IR datasets generated via the three methods to be described in Section 3. 
To fuse the results of our RGB and IR deep learning networks, we propose a lightweight IAN  shown in Fig. \ref{fig:IAN}.
This is motivated by the fact that RGB images result in better performance at daytime due to their higher spatial resolution and multiplicities of channels, namely R, G, and B, whereas at night-time the single channel IR image outperforms the visible light. 
The input to IAN is an RGB/IR image pair and the output consists of the RGB and IR weights which are used to select the appropriate detector.
It is trained in coordination with the RGB and IR detectors, in that it selects the higher performing detector rather than predicting illumination conditions.
This is because illumination may not be the only factor that affects accuracy.
As seen in Fig. \ref{fig:IAN}, the IAN is comprised of 7 layers with the first 5 being convolution + max pooling layers and the last 2 being fully connected layers. 

\begin{figure*}[htp]
    \centering
    \includegraphics[width=\linewidth]{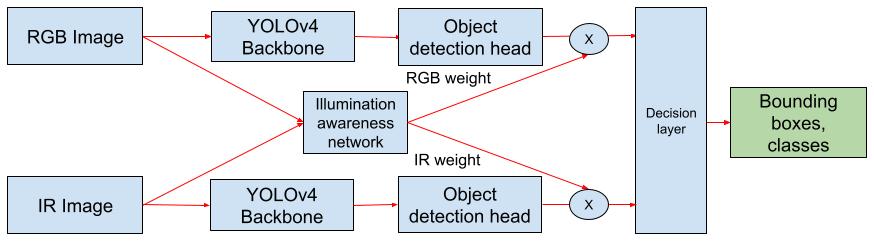}
    \caption{Block diagram of our proposed system comprising of an RGB and IR YOLOv4 detector and an IAN network fusing the two.  RGB/IR images are passed to their respective detectors and to the IAN, generating object detection results and confidence weights. The decision layer decides which model to trust and outputs the result.}
    \label{fig:late-fusion}
\end{figure*}

To take full advantages of RGB and IR detectors, we propose a late fusion structure as seen in Fig. \ref{fig:late-fusion}. 
It is comprised of two YOLOv4 detectors and an IAN network.
Paired RGB/IR images are passed to their respective detectors and to the IAN, generating object detection results and confidence weights.
The decision layer decides which model to trust and outputs the result.

\section{3. Synthetic IR data generation}

To train a deep learning object detection model, we need large amounts of labelled, paired RGB/IR data. 
Paired RGB/IR drone image datasets are rare, and even if they exist, they have very few examples, and are not labelled. 
Also, in the absence of parried RGB images, it is not easy for human operators to accurately label IR images. 
In this section, we describe three methodologies for creating labelled IR images, which are paired with labelled RGB images to provide as input to our system shown in Fig. \ref{fig:late-fusion}. 
In this paper, we are primarily interested in two classes of objects: pedestrian and cars. 
Our first approach, described in Section 3.1 is based on CycleGAN, the second one described in Section 3.2 is based on AIRSim and the third one in Section 3.3. is a combination of the two. 

\label{subsec:sync}
\subsection{3.1 CycleGAN with Mask R-CNN Segmentation }
\begin{figure}[]
    \centering
    \includegraphics[width=\linewidth]{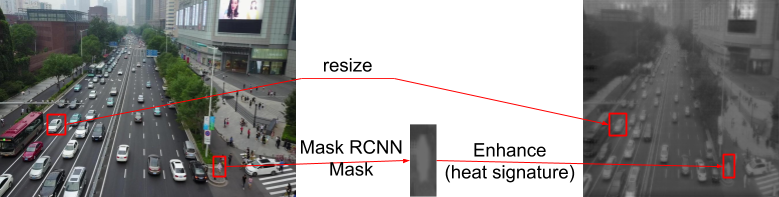}
    \vspace{0.1cm}
    \caption{Generating images via CycleGAN and Mask R-CNN; the RGB image is resized to $1024\times1024$ from $3326\times1871$ and passed into the CycleGAN model trained with unpaired RGB/IR images from the same data distribution; objects of interest , i.e. humans and cars are enhanced via Mask R-CNN mask semantic segmentation and heat signature modification.}
    \label{fig:c-m-archi}
\end{figure}

\begin{figure}[]
\vspace{-1.5cm}
        \centering
        \begin{minipage}{\linewidth}
            \centering
            \includegraphics[width=\linewidth]{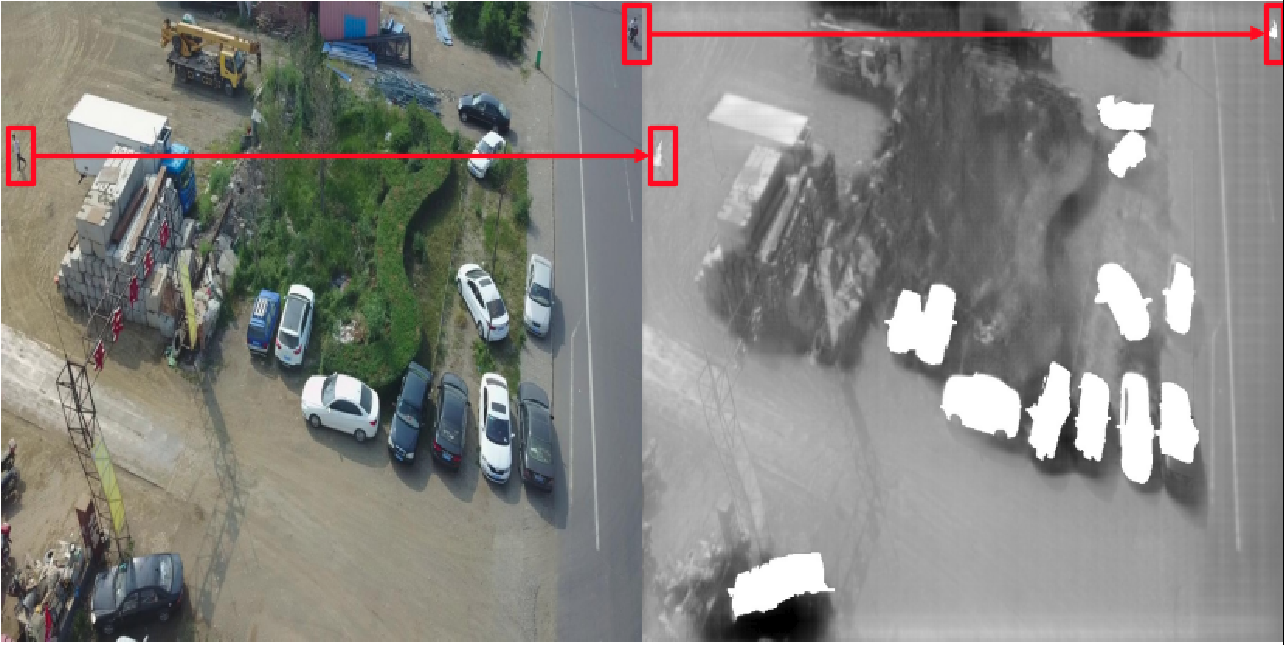}
            \subcaption{}
                    \vspace{0.5cm}
            \label{fig:cmd1}
        \end{minipage}
  \vspace{0.2cm}
        \begin{minipage}{\linewidth}
            \centering
            \includegraphics[width=\linewidth]{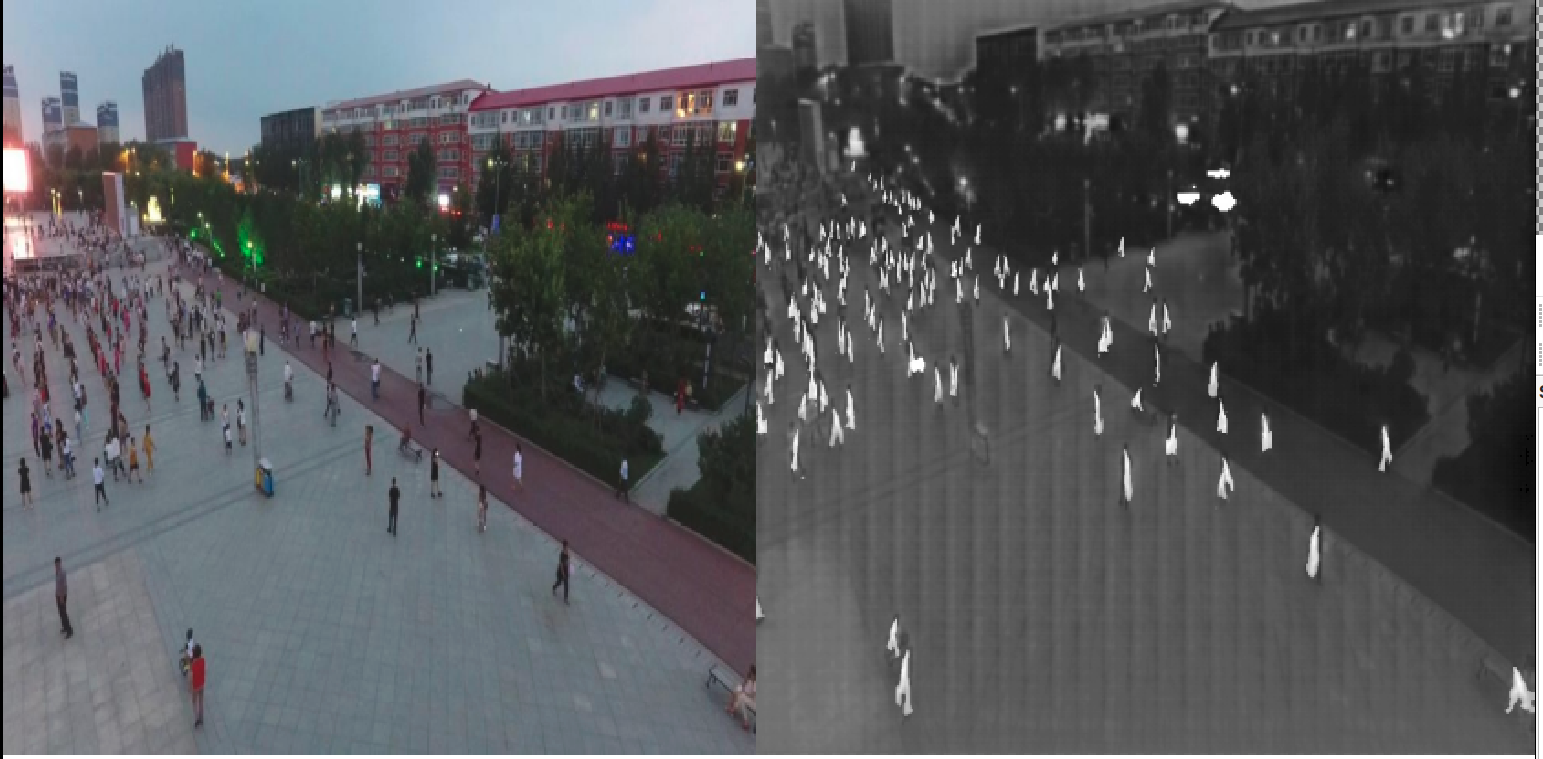}
            \subcaption{}
             \vspace{0.3cm}
            \label{fig:cmd2}
        \end{minipage}
              \vspace{0.2cm}
        \begin{minipage}{\linewidth}
            \centering
            \includegraphics[width=\linewidth]{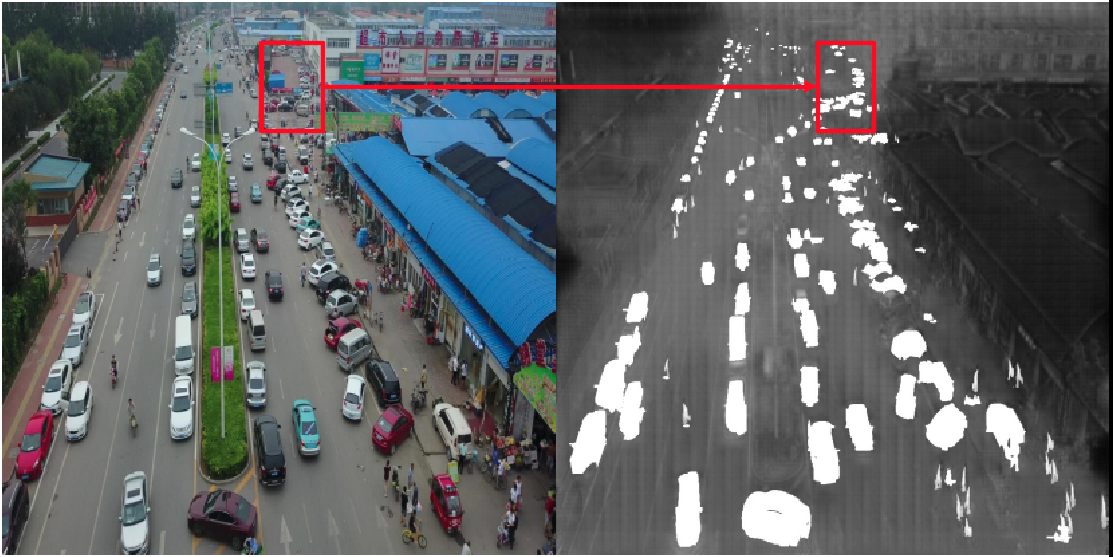}
            \subcaption{}
            \label{fig:cmd3}
        \end{minipage}
         \vspace{0.3cm}
        \caption{Examples of CycleGAN \& Mask R-CNN Synthetic IR generation: (a) Mask R-CNN does not perform well on small humans; (b) Simply tuning the segmented out areas is not natural; (c) No distinction between objects close together.}
        \label{fig:cmd}
\end{figure}

To generate the IR images from their RGB counter parts, we train a CycleGAN \cite{CycleGAN2017} using unpaired RGB/IR images data captured with our own drone, as well as existing unlabelled non-paired images \cite{iricra2014} to generate an RGB-to-IR style adapter. 
To create labelled IR images, we apply an accurate off-the-shelf object detector to the RGB images that are then fed to the CycleGAN. 
We then translate the bounding boxes from the RGB images to the IR images as shown in Fig. \ref{fig:c-m-archi}. 
Such RGB detectors with mAP of larger than 90\% are readily available in the literature \cite{he2017mask,shah2017airsim}.
To create more realistic IR imagery, we also apply Mask R-CNN \cite{he2017mask} to the bounding boxes of detected objects in RGB images to semantically segment out the humans and cars in order to enhance their heat signature.
In doing so, we make them appear to be warmer than the background environment by adjusting the value of the segmented pixels, as shown in Fig. \ref{fig:c-m-archi}.  
The RGB image is resized from $3326\times1871$ to $1024\times1024$ before it is passed onto the CycleGAN model, which is trained with unpaired RGB/IR images from the same data distribution of urban environments.
Example synthetic IR data generated using this approach, along with their RGB counterparts can be found in Fig. \ref{fig:cmd1}-\ref{fig:cmd3}. 
Effectively, the CycleGAN enables us to take advantage of existing RGB drone image datasets such as VisDrone in order to create a large number of synthetic IR imagery using the trained CycleGAN. 
That said, this method has several drawbacks: first to create IR realistic images, the CycleGAN requires a large number of input examples, both RGB and IR; while there is an abundance of drone RGB images, the same cannot be said with IR; second, small number of pixels associated with pedestrians and cars in the drone imagery adversely affects the performance of Mask R-CNN instance segmentation as seen in Fig. \ref{fig:cmd1}.
Applying classical segmentation schemes such as GrabCut \cite{grabcut} does not result in better instance segmentation either due to the small number of pixels in the bounding box associated with the objects of interest.
Thirdly, simple tuning of the heat signature via adjustment of the pixel values of the object looks unnatural, as shown in Fig. 4b.
Finally, Mask R-CNN produces unacceptable segmentation masks in situations with numerous objects in proximity of each other, as shown in the example in  Fig. \ref{fig:cmd3}. 

\subsection{3.2 Simulation Rendering of Environment}
\begin{figure}[]
    \centering
    \includegraphics[width=\linewidth]{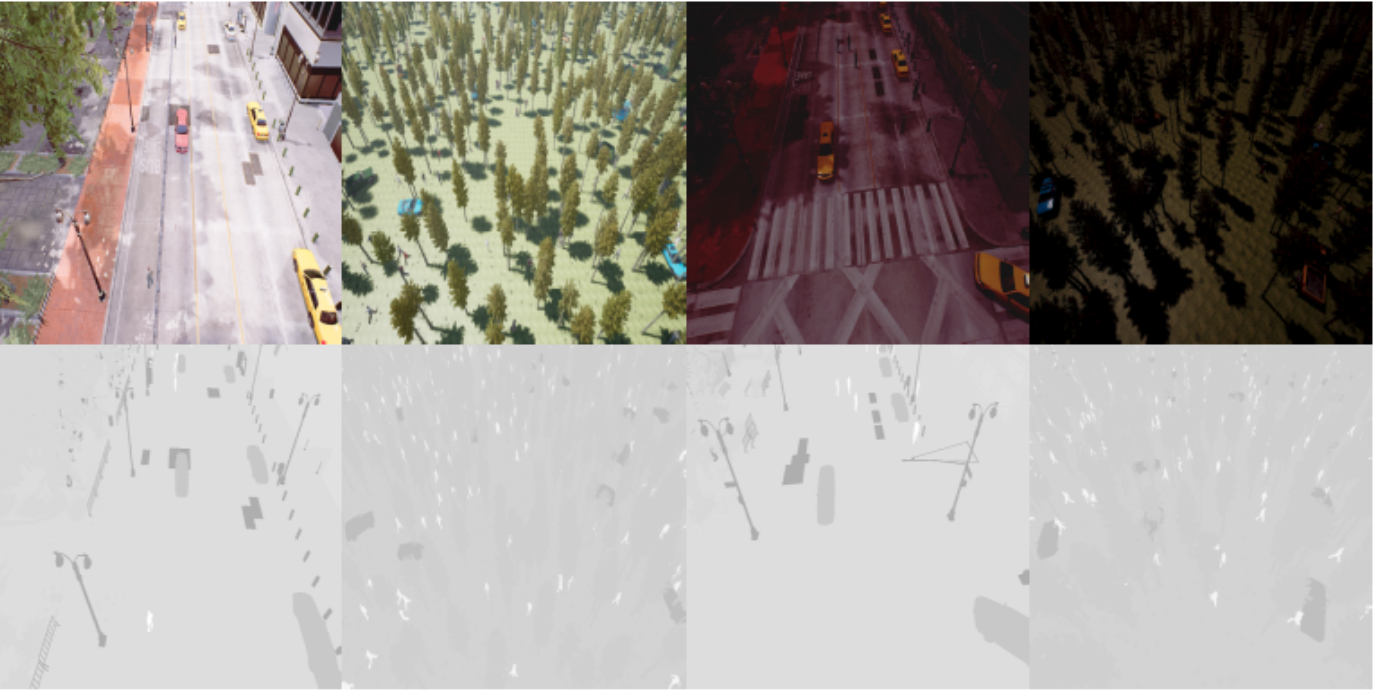}
    \caption{Examples of using the AIRSim simulator to capture images.}
    \label{fig:sd1}
\vspace{0.2cm}
        \centering
        \begin{minipage}{\linewidth}
            \centering
            \includegraphics[width=0.65\linewidth,height=0.5\linewidth]{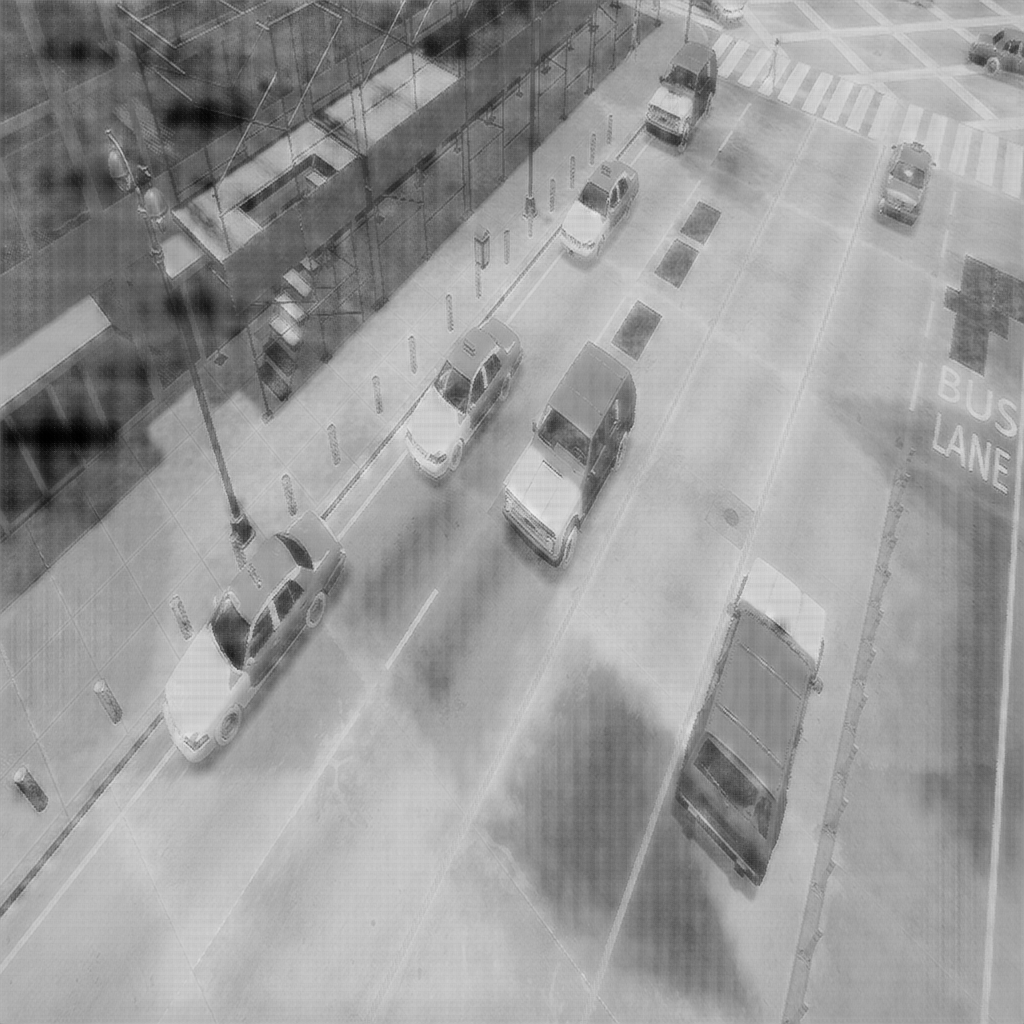}
            \subcaption{}
            \label{fig:sc-2}
        \end{minipage}
        \begin{minipage}{\linewidth}
  \vspace{0.3cm}
            \centering
            \includegraphics[width=0.65\linewidth]{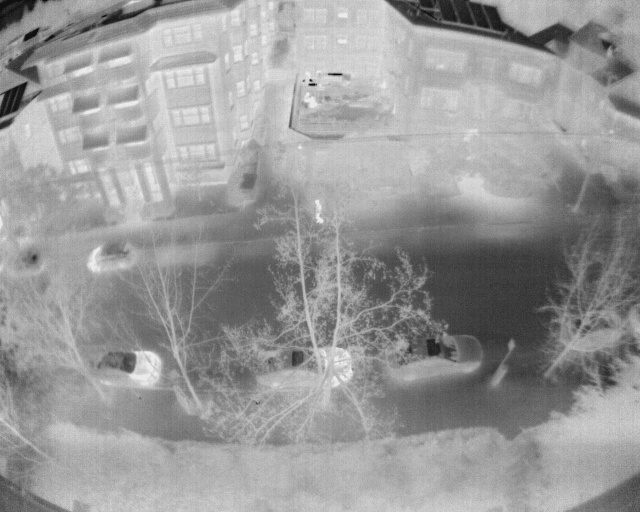}
            \subcaption{}
            \label{fig:sc-1}
        \end{minipage}
          \vspace{0.3cm}
        \caption{Comparison of simulation and CycleGAN generated IR images with actual IR images captured by a thermal camera; (a) synthetic IR images generated by passing simulated RGB images into CycleGAN trained with real world data, plus additional temperature adjustments; for example, engine hoods of the cards have been artificially made "hot" ; (b) Actual IR image captured with a thermal camera on situated on a rooftop of a building in Berkeley, California.}
        \vspace{0.3cm}
                \centering
        \begin{minipage}{0.49\linewidth}
            \centering
            \adjincludegraphics[width=\linewidth,trim={0 0 {.5\width} 0},clip]{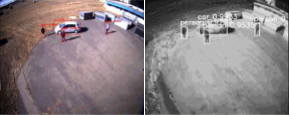}
            \subcaption{}
                    \vspace{0.3cm}
        \end{minipage}
        \begin{minipage}{0.49\linewidth}
            \centering
            \adjincludegraphics[width=\linewidth,trim={{.5\width} 0 0 0},clip]{Figs/header.png}
            \subcaption{}
                    \vspace{0.2cm}
        \end{minipage}
    \caption{Detection results from the RGB/IR models onboard the Nvidia Xavier drone, during the experiment carried out in Richmond Field Station, Berkeley, California. (a) RGB output with red bounding boxes; (b) IR output with white bounding boxes. Both models detect the three people and one car in the scene successfully.}
    \label{fig:my_label}
\end{figure}

To overcome some of the above-mentioned problems in the CycleGAN \& Mask R-CNN approach, we seek new methods to generate labelled paired IR/RGB images.
As simulation in Unreal engine can create semi-realistic environments of our choosing, we leverage the power of AIRSim \cite{shah2017airsim}, a high-fidelity physical and visual drone flight simulator, to synthetically generate paired IR/RGB images for training our deep learning networks of Fig. \ref{fig:late-fusion}.
We model the temperature distribution of the humans, vehicles and background following the model presented in \cite{bondi2018airsim}, pre-assigning temperatures to different categories of objects during different day/night periods.
We create both forest and city virtual environments, placing different models of humans and cars in them with randomized positions and orientations, then perform data collection with a camera angled downwards at 45 degrees, with example images shown in Fig. \ref{fig:sd1}.
With this simulation approach, we have unparalleled control of the environment and the lighting, without needing to adhere to the priors in RGB datasets such as VisDrone which are often limited to one type of a scene e.g. urban environments. 
Even though the resulting synthetic images from AIRSim are realistic, they do not completely bridge the sim-to-real gap between the simulation generated IR imagery and actual IR imagery.

\subsection{3.3 Combining CycleGAN with Simulation Based Rendering} 
To bridge the sim-to real gap, we opt to combine CycleGAN with AIRSim as follows. 
We replace 50\% of the original RGB training dataset used for the CycleGAN-only method with the AIRSim generated RGB images, and keep the remaining 50\% RGB training images from VisDrone intact. 
We also keep the IR training set unchanged with respect to the CycleGAN-only method. 
During inference, we pass AIRSim rendered RGB images through the CycleGAN. 
This approach enables us to keep the original style transfer quality of the real-world images in CycleGAN, while also ensuring that the synthetic RGB images are successfully style-transferred to their IR counterparts. 
In this case, since the RGB images emanate from AIRSim simulations in inference, the IR labels are perfectly accurate and do not suffer from imperfections of R-CNN segmentation due to small object sizes. 
Furthermore, as we have full control over the lighting and environment, we can create more realistic  daytime and night-time IR images, and introduce thermal variation on the surface of objects, as seen in Fig. \ref{fig:sc-2}. 
In this picture, we make the engine hoods of the cars "hot" so that it looks similar to an actual IR image shown in Fig. \ref{fig:sc-1}. 
Again, the simulation environment provides us with pixel perfect segmentation of the objects and hence allows us to manipulate the exact pixels in the object for better realism.

%% file: experiments.tex
\begin{table*}[h]
\centering
\begin{tabular}{|c|ccc|ccc|cccc|c|}
\hline
Class\textbackslash{}Detector & \multicolumn{3}{c|}{RGB} & \multicolumn{3}{c|}{IR} & \multicolumn{4}{c|}{Fusion} & IAN \\ \hline
Setting & \multicolumn{1}{c|}{Day} & \multicolumn{1}{c|}{Night} & All & \multicolumn{1}{c|}{Day} & \multicolumn{1}{c|}{Night} & All & \multicolumn{1}{c|}{Day} & \multicolumn{1}{c|}{Night} & \multicolumn{1}{c|}{All} & \multicolumn{1}{l|}{Oracle} & \multirow{2}{*}{Precision} \\ \cline{1-11}
Car & \multicolumn{1}{c|}{0.86} & \multicolumn{1}{c|}{0.80} & 0.83 & \multicolumn{1}{c|}{0.84} & \multicolumn{1}{c|}{0.70} & 0.77 & \multicolumn{1}{c|}{0.85} & \multicolumn{1}{c|}{0.77} & \multicolumn{1}{c|}{0.81} & 0.86 &  \\ \hline
People & \multicolumn{1}{c|}{0.55} & \multicolumn{1}{c|}{0.43} & 0.49 & \multicolumn{1}{c|}{0.09} & \multicolumn{1}{c|}{0.01} & 0.05 & \multicolumn{1}{c|}{0.45} & \multicolumn{1}{c|}{0.34} & \multicolumn{1}{c|}{0.40} & 0.54 & \multirow{2}{*}{0.75} \\ \cline{1-11}
mAP@0.5 IOU & \multicolumn{1}{c|}{0.71} & \multicolumn{1}{c|}{0.62} & \textbf{0.66} & \multicolumn{1}{c|}{0.46} & \multicolumn{1}{c|}{0.36} & \textbf{0.41} & \multicolumn{1}{c|}{0.65} & \multicolumn{1}{c|}{0.56} & \multicolumn{1}{c|}{\textbf{0.61}} & 0.70 &  \\ \hline
\end{tabular}
\caption{RGB/IR Fusion with CycleGAN}
\label{tab:1}
\vspace{0.4cm}
\centering
\begin{tabular}{|c|ccc|ccc|cccc|c|}
\hline
Class\textbackslash{}Detector & \multicolumn{3}{c|}{RGB} & \multicolumn{3}{c|}{IR} & \multicolumn{4}{c|}{Fusion} & IAN \\ \hline
Setting & \multicolumn{1}{c|}{Day} & \multicolumn{1}{c|}{Night} & All & \multicolumn{1}{c|}{Day} & \multicolumn{1}{c|}{Night} & All & \multicolumn{1}{c|}{Day} & \multicolumn{1}{c|}{Night} & \multicolumn{1}{c|}{All} & Oracle & \multirow{2}{*}{Precision} \\ \cline{1-11}
Car & \multicolumn{1}{c|}{0.98} & \multicolumn{1}{c|}{0.92} & 0.95 & \multicolumn{1}{c|}{0.72} & \multicolumn{1}{c|}{0.78} & 0.75 & \multicolumn{1}{c|}{0.86} & \multicolumn{1}{c|}{0.82} & \multicolumn{1}{c|}{0.84} & 0.97 &  \\ \hline
People & \multicolumn{1}{c|}{0.65} & \multicolumn{1}{c|}{0.57} & 0.61 & \multicolumn{1}{c|}{0.81} & \multicolumn{1}{c|}{0.89} & 0.85 & \multicolumn{1}{c|}{0.90} & \multicolumn{1}{c|}{0.80} & \multicolumn{1}{c|}{0.85} & 0.87 & \multirow{2}{*}{0.83} \\ \cline{1-11}
mAP@0.5 IOU & \multicolumn{1}{c|}{0.82} & \multicolumn{1}{c|}{0.74} & \textbf{0.78} & \multicolumn{1}{c|}{0.76} & \multicolumn{1}{c|}{0.84} & \textbf{0.80} & \multicolumn{1}{c|}{0.88} & \multicolumn{1}{c|}{0.81} & \multicolumn{1}{c|}{\textbf{0.85}} & 0.92 &  \\ \hline
\end{tabular}
\caption{RGB/IR Fusion with Simulation}
\label{tab:2}
\vspace{0.4cm}
\centering
\begin{tabular}{|c|ccc|ccc|cccc|c|}
\hline
Class\textbackslash{}Detector & \multicolumn{3}{c|}{RGB} & \multicolumn{3}{c|}{IR} & \multicolumn{4}{c|}{Fusion} & IAN \\ \hline
Setting & \multicolumn{1}{c|}{Day} & \multicolumn{1}{c|}{Night} & All & \multicolumn{1}{c|}{Day} & \multicolumn{1}{c|}{Night} & All & \multicolumn{1}{c|}{Day} & \multicolumn{1}{c|}{Night} & \multicolumn{1}{c|}{All} & \multicolumn{1}{l|}{Oracle} & \multirow{2}{*}{Precision} \\ \cline{1-11}
Car & \multicolumn{1}{c|}{0.98} & \multicolumn{1}{c|}{0.92} & 0.95 & \multicolumn{1}{c|}{0.99} & \multicolumn{1}{c|}{0.99} & 0.99 & \multicolumn{1}{c|}{0.98} & \multicolumn{1}{c|}{0.97} & \multicolumn{1}{c|}{0.97} & 0.99 &  \\ \hline
People & \multicolumn{1}{c|}{0.65} & \multicolumn{1}{c|}{0.57} & 0.61 & \multicolumn{1}{c|}{0.94} & \multicolumn{1}{c|}{0.98} & 0.96 & \multicolumn{1}{c|}{0.91} & \multicolumn{1}{c|}{0.95} & \multicolumn{1}{c|}{0.93} & 0.97 & \multirow{2}{*}{0.96} \\ \cline{1-11}
mAP@0.5 IOU & \multicolumn{1}{c|}{0.82} & \multicolumn{1}{c|}{0.74} & \textbf{0.78} & \multicolumn{1}{c|}{0.97} & \multicolumn{1}{c|}{0.99} & \textbf{0.98} & \multicolumn{1}{c|}{0.94} & \multicolumn{1}{c|}{0.96} & \multicolumn{1}{c|}{\textbf{0.95}} & 0.98 &  \\ \hline
\end{tabular}
\caption{RGB/IR Fusion with Simulation \& CycleGAN}
\label{tab:3}
\end{table*}
\begin{table*}[]
\centering
\begin{tabular}{|c|c|c|}
\hline
Method\textbackslash{}Dataset & RFS-day & Hearst-night \\ \hline
CycleGAN & \begin{tabular}[c]{@{}c@{}}\textbf{car AP = 27.95\%}\\    \\ person AP =   0.06\%\\    \\ mAP = 14.01\%\end{tabular} & \begin{tabular}[c]{@{}c@{}}car AP = 0.22\%\\    \\ person AP =   0.00\%\\    \\ mAP = 0.11\%\end{tabular} \\ \hline
AIRSim Simulation & \begin{tabular}[c]{@{}c@{}}car AP = 0.50\%\\    \\ person AP = 0.23\%\\    \\ mAP = 0.37\%\end{tabular} & \begin{tabular}[c]{@{}c@{}}car AP = 0.01\% \\    \\ person AP =   0.42\% \\    \\ mAP = 0.22\%\end{tabular} \\ \hline
CycleGAN \& AIRSim & \begin{tabular}[c]{@{}c@{}}car AP = 0.00\% \\    \\ person AP = 5\% \\    \\ mAP = 2.5\%\end{tabular} & \begin{tabular}[c]{@{}c@{}}car AP = 0.00\% \\    \\ person AP = 2\% \\    \\ mAP = 1\%\end{tabular} \\ \hline
RGB detector with IR as grayscale image input & \begin{tabular}[c]{@{}c@{}}\textbf{car AP = 33\%} \\    \\ person AP = 6\% \\    \\ mAP = 19.5\%\end{tabular} & \begin{tabular}[c]{@{}c@{}}\textbf{car AP = 14.91\%}   \\    \\ person AP =   0.03\% \\    \\ mAP = 7.47\%\end{tabular} \\ \hline
\end{tabular}
\caption{Comparison of various synthetic IR generation methods tested on real-world images of both day and night settings.}
\label{tab:4}
\end{table*}
\section{4. Experiments}
We test our trained network on both real-world and synthetically generated RGB/IR paired datasets for the three methods described in Section 3. 
For synthetic datasets, the distribution between night time and day time is 50/50.
Further, they are split 80/20 into training and testing respectively. 
For the VisDrone dataset used in CycleGAN, the test set images are not used during training and are held out.
For AIRSim-based methods, the test sets are generated with different distributions of human and car models. The hyper-parameters are directly tuned on the training set.
All datasets are run on RGB-only, IR-only and fusion models separately, with the results shown in Tables \ref{tab:1}-\ref{tab:3}.
In general, there are two factors contributing to the performance of fusion: the performance of IAN as measured by mAP and the degree of redundancy between IR and RGB images, i.e. how well they complement each other.

Table \ref{tab:1} shows the results for the CycleGAN approach with RGB data used for training and testing from VisDrone dataset.
As seen, the RGB-only model outperforms the IR-only model.
One possible explanation is that for the VisDrone dataset, we leverage the existing annotations of the RGB images to annotate the generated IR images.
Specifically, we copy the car bounding boxes from the RGB images to their IR counterparts, and pass the detected humans through the aforementioned thermal signature augmentation pipeline.
Thus, by definition, in this approach, IR could never detect any more objects than RGB can or outperform RGB since the IR annotations are a subset of the RGB annotations.
In practice, this is not the case since IR images captured with actual thermal cameras can reveal objects in dark while RGB cannot. In effect the CycleGAN approach does not physically model the fact that at night times, objects are detectable in IR images, but not in RGB images.
It is noted that the fusion model does worse than IR and RGB standalone models. This is because the RGB outperforms IR by more than 20\% and the IAN is only 75\% accurate, leading to a performance drop in the fusion approach.

Table \ref{tab:2} shows the results for AIRSim simulated RGB/IR imagery. 
Unlike Table 1 where RGB performance is 20\% higher than IR, in Table 2, the gap between RGB and IR is only 2\%.
As seen, overall fusion mAP is higher than RGB-only or IR-only methods.
This can be explained by the fact that RGB and IR detectors are complimenting each other at day and night time. Specifically, unlike RGB detector which performs poorly at night, the IR detector has specifically been trained to detect objects in dark. Here the IAN can choose the better performing detector with accuracy of 83\%.

Table \ref{tab:3} shows the performance of the combined CycleGAN and Simulation based rendering. 
Recall that in this case, the labeled synthesized IR imagery is as accurate as labeled RGB imagery because we use simulated RGB images as input to the CycleGAN for both daytime and night-time images.
Consequently, as seen in Table \ref{tab:3} the performance of IR-only is much improved compared to Table 1. 
In addition, in Table \ref{tab:3}, IR outperforms RGB by about 20\% since no information is lost in the IR images due to illumination conditions.
After fusion with RGB, due to the still imperfect IAN , the fusion performance is slightly  worse than IR.

We also test our IR model trained with various methods on actual IR images captured with an IR camera as shown in Table \ref{tab:4}. 
For the real-world dataset, we collected and annotated 110 daytime images at Richmond Field Station (RFS-day) and 160 night-time images at Etcheverry Building overlooking Hearst Avenue (Hearst-night).
The IR camera used is the FLIR Boson 640 with a resolution at 640x512. 
There are 403 instances of cars and 346 instances of people in the RFS-day dataset and 496 instances of cars and 310 instances of people in the Hearst-night dataset. 
Neither set were used in training. 
As seen, the performance of all three synthetic IR generation methods is poor, and significantly worse than their test results on synthetic data in Tables \ref{tab:1}-\ref{tab:3}. 
The fourth row of Table \ref{tab:4} corresponds to an IR detector which simply converts IR image into grayscale image and applies it as input to the RGB-only model.
As seen, this rather simple method, outperforms all the other three synthetic IR generation methods. The only synthetic method that comes close to this simple “grayscale” IR detector is daytime CycleGAN detection of cars at 27.95\% accuracy. 
Also note that combination of CycleGAN plus AIRSim is superior to AIRSim-only, confirming that the sim-to-real gap is somewhat narrowed. 

We implemented our fused detection model on an Nvidia Xavier shared with other sensor processing.
The processing time for our fused detection with both the RGB and the grayscale IR model is 28 milliseconds per RGB/IR pair.
An example of such fused detection on an actual image captured with our drone during the experiment carried out in Richmond Field Station, Richmond, California is shown in Fig. \ref{fig:my_label}.
Specifically, the left and right images show the RGB  and IR output with red and white bounding boxes respectively. 
Both models detect the three people and one car in the scene in Fig. \ref{fig:my_label} successfully. The performance of the fusion models can be seen in Tables \ref{tab:5} and \ref{tab:6}. 
As expected, for both daytime and night time datasets, RGB does better than IR, and fusion does better than either one. 
Furthermore, the overall performance is considerably better at daytime than night time. This underlines the need for more realistic IR data for training, particularly at night time. 

\begin{table}[h]
\centering
\resizebox{\columnwidth}{!}{%
\begin{tabular}{|c|c|c|c|}
\hline
Class\textbackslash{}Detector & RGB detector & Grayscale IR   detector & Fusion \\ \hline
Car & 0.78 & 0.33 & 0.79 \\ \hline
People & 0.84 & 0.06 & 0.83 \\ \hline
mAP@0.5 IOU & 0.81 & 0.20 & 0.81 \\ \hline
\end{tabular}
}
\caption{Fusion performance on RFS-day dataset}
\label{tab:5}
\vspace{0.3cm}
\centering
\resizebox{\columnwidth}{!}{%
\begin{tabular}{|c|c|c|c|}
\hline
Class\textbackslash{}Detector & RGB detector & Grayscale IR   detector & Fusion \\ \hline
Car & 0.25 & 0.15 & 0.27 \\ \hline
People & 0.01 & 0.0003 & 0.01 \\ \hline
mAP@0.5 IOU & 0.13 & 0.08 & 0.14 \\ \hline
\end{tabular}
}
\caption{Fusion performance on Hearst-night dataset}
\label{tab:6}
\end{table}

%% file: conclusion.tex
\section{5. Conclusion}
In this paper, we presented an RGB/IR fusion detection framework. 
Future work involves developing methods to improve the performance of IR synthetic image generation in such a way that the resulting models work better on actual IR object detection models, i.e. closing the sim-to-real gap. 
Synthesizing realistic IR images is particularly challenging at night.
Few-shot or one-shot learning could also be worthwhile to investigate to alleviate shortage of night time drone imagery. 
\section{6. Acknowledgements}
We thank the Army Research Lab and the Microsoft Humanitarian AI fund for funding this project.

%% file: root.bbl
\small
\begin{thebibliography}{9}
\bibitem{yolov4}Bochkovskiy, Alexey, Chien-Yao Wang, and Hong-Yuan Mark Liao, Yolov4: Optimal speed and accuracy of object detection, arXiv preprint arXiv:2004.10934 (2020).
\bibitem{zhu2018vision}Zhu, Pengfei, et al., Vision meets drones: A challenge, arXiv preprint arXiv:1804.07437 (2018).
\bibitem{iricra2014}Thermal infrared dataset, [Online]. Available:  \url{https://projects.asl.ethz.ch/datasets/doku.php?id=ir:iricra2014}
\bibitem{leykin2007thermal}Leykin, Alex, Yang Ran, and Riad Hammoud, Thermal-visible video fusion for moving target tracking and pedestrian classification, Proc. CVPR, pg. 1-8. (2007).
\bibitem{takumi2017multispectral}Takumi, Karasawa, et al., Multispectral object detection for autonomous vehicles, Proc. Thematic Workshops, ACM Multimedia, pg. 35–43. (2017).
\bibitem{coaxials-2019}Okazawa, Atsuro, Tomoyuki Takahata, and Tatsuya Harada, Simultaneous transparent and non-transparent object segmentation with multispectral scenes, Proc. IROS, pg. 4977-4984. (2019).
\bibitem{multispectral} Chen, Yunfan, and Hyunchul Shin, Multispectral image fusion based pedestrian detection using a multilayer fused deconvolutional single-shot detector, J. Opt. Soc. Am. A 37.5, 768–779 (2020).
\bibitem{guan-illumination-aware}Guan, Dayan, et al., Fusion of multispectral data through illumination-aware deep neural networks for pedestrian detection,  J. Information Fusion, 50, 148-157 (2019).
\bibitem{li-illumination-aware}Li, Chengyang, et al., Illumination-aware faster R-CNN for robust multispectral pedestrian detection, J. Pattern Recognition, 85, 161-171(2019).
\bibitem{leakage}Li, Anqi, et al., Rgb-thermal fusion network for leakage detection of crude oil transmission pipes, Proc. ROBIO, pg. 883-888. (2019).
\bibitem{spectral-edge-fusion}French, Geoff, Graham Finlayson, and Michal Mackiewicz,  Multi-spectral pedestrian detection via image fusion and deep neural networks, Proc. CIC, pg. 176-181. (2018).
\bibitem{FasterR-CNN}Ren, Shaoqing, et al., Faster r-cnn: Towards real-time object detection with region proposal networks, Proc. NeurIPS, pg. 91-99. (2015).
\bibitem{RetinaNet}Lin, Tsung-Yi, et al. "Focal loss for dense object detection." Proc. ICCV, pg. 2980-2988. (2017).
\bibitem{SSD}Liu, Wei, et al., Ssd: Single shot multibox detector, Proc. ECCV, pg. 21-37. (2016).
\bibitem{yolov3}Farhadi, Ali, and Joseph Redmon, Yolov3: An incremental improvement, Proc. CVPR, pg. 1804-02767. (2018).
\bibitem{tan2020efficientdet}Tan, Mingxing, Ruoming Pang, and Quoc V. Le., Efficientdet: Scalable and efficient object detection, Proc. CVPR, pg. 10781-10790. (2020).
\bibitem{lin2015microsoft-coco}Lin, Tsung-Yi, et al., Microsoft coco: Common objects in context, Proc. ECCV, pg. 740-755. (2014).
\bibitem{zhu2020vision}Zhu, Pengfei, et al., Vision meets drones: Past, present and future, arXiv preprint arXiv:2001.06303. (2020).
\bibitem{CycleGAN2017}Zhu, Jun-Yan, et al., Unpaired image-to-image translation using cycle-consistent adversarial networks, Proc. ICCV, pg 2223-2232. (2017).
\bibitem{he2017mask}He, Kaiming, et al., Mask r-cnn, Proc. ICCV, pg 2961-2969. (2017).
\bibitem{shah2017airsim}Shah, Shital, et al., Airsim: High-fidelity visual and physical simulation for autonomous vehicles, Proc. FSR, pg. 621-635. (2018).
\bibitem{bondi2018airsim} Bondi, Elizabeth, et al., Airsim-w: A simulation environment for wildlife conservation with uavs, Proc. ACM COMPASS, pg. 1-12. (2018).
\bibitem{unreal} Epic Games, Unreal Engine [Internet]. Available from: \url{https://www.unrealengine.com}.(2019).
\bibitem{spp} He, Kaiming, et al. "Spatial pyramid pooling in deep convolutional networks for visual recognition." IEEE Trans. Patt. An. \& Mach. Intelli. 37.9, pg. 1904-1916.  (2015)
\bibitem{grabcut} Rother, Carsten, Vladimir Kolmogorov, and Andrew Blake. "'GrabCut' interactive foreground extraction using iterated graph cuts" ACM Trans. Graphics 23.3 pg. 309-314. (2004).
\end{thebibliography}
